\documentclass[runningheads]{llncs}

\usepackage{graphicx}
\usepackage{amsmath}
\usepackage[ruled,vlined]{algorithm2e}
\DeclareMathOperator*{\argmin}{argmin}

\usepackage{amssymb}
\usepackage{subcaption} 

\renewcommand{\v}[1]{{\boldsymbol{\mathbf{#1}}}}
\newcommand\state{\v{x}}
\newcommand\obs{\v{o}}
\newcommand\action{\v{a}}

\begin{document}
	
	\title{Robot Localization and Navigation through Predictive Processing using LiDAR\thanks{2nd International Workshop on Active Inference IWAI2021, European Conference on Machine Learning (ECML/PCKDD 2021)}}
	\author{D. Burghardt\inst{1} \and
		P. Lanillos\inst{2}}
	\institute{Radboud University, Houtlaan 4, 6525 XZ Nijmegen, NL \and
		Donders Institute for Brain, Cognition and Behaviour, Department of Artificial Intelligence, Radboud University, Nijmegen, NL}
	
	\titlerunning{Localization and navigation through predictive processing using LiDAR}
	\maketitle              
	\begin{abstract}
		Knowing the position of the robot in the world is crucial for navigation. Nowadays, Bayesian filters, such as Kalman and particle-based, are standard approaches in mobile robotics. Recently, end-to-end learning has allowed for scaling-up to high-dimensional inputs and improved generalization. However, there are still limitations to providing reliable laser navigation. Here we show a proof-of-concept of the predictive processing-inspired approach to perception applied for localization and navigation using laser sensors, without the need for odometry. We learn the generative model of the laser through self-supervised learning and perform both online state-estimation and navigation through stochastic gradient descent on the variational free-energy bound. We evaluated the algorithm on a mobile robot (TIAGo Base) with a laser sensor (SICK) in Gazebo. Results showed improved state-estimation performance when comparing to a state-of-the-art particle filter in the absence of odometry. Furthermore, conversely to standard Bayesian estimation approaches our method also enables the robot to navigate when providing the desired goal by inferring the actions that minimize the prediction error.
		
	\end{abstract}
	
	\section{Introduction}
	Localization algorithms are part of our daily life and core for robotics. Recursive Bayesian estimation composes the current state-of-art in the field and has been essential for the development of localization, mapping, navigation and searching applications~\cite{thrun_2007:SLAM,lanillos2013minimum}. Bayesian filters~\cite{bayes_filtering}, e.g., Kalman and particle filters, are able to estimate the state of a system from noisy sensor observations formalized as a hidden Markov model. These approaches are useful also in the case of non-linear modeled systems and out-of-sequence measurements~\cite{besada2012localization}. The particle filter (PF) is an approximate Bayesian method that tractably computes the posterior distribution of the state of any system given the observations. The state distribution is represented by individual particles, which are evaluated and weighted recursively. Particles with higher probability get bigger weights and are re-sampled into more particles in its neighborhood, whereas particles with smaller weights get fewer new samples close to them \cite{particle_filter}.
	
	In recent years, novel approaches based on deep neural networks, have been proposed to improve localization using high-dimensional inputs. Regression solutions, for instance, compute the absolute position of the system from only visual information~\cite{kendall2015posenet}. However, these methods have lower accuracy than previous approaches that exploit prior information, such as geometry~\cite{sattler2019understanding}. In particular, LiDAR-based Navigation with representation learning (e.g., using autoencoders) and reinforcement learning has shown downgraded performance in navigation tasks~\cite{gebauer2021pitfall}. Alternatively to LiDAR-based approaches, \cite{catal2020} and \cite{pixel_ai} showed how to apply deep active inference using camera images in robot navigation and humanoid upper-body reaching respectively.
	
\begin{figure}
		\centerline{\resizebox{12cm}{!}{\includegraphics{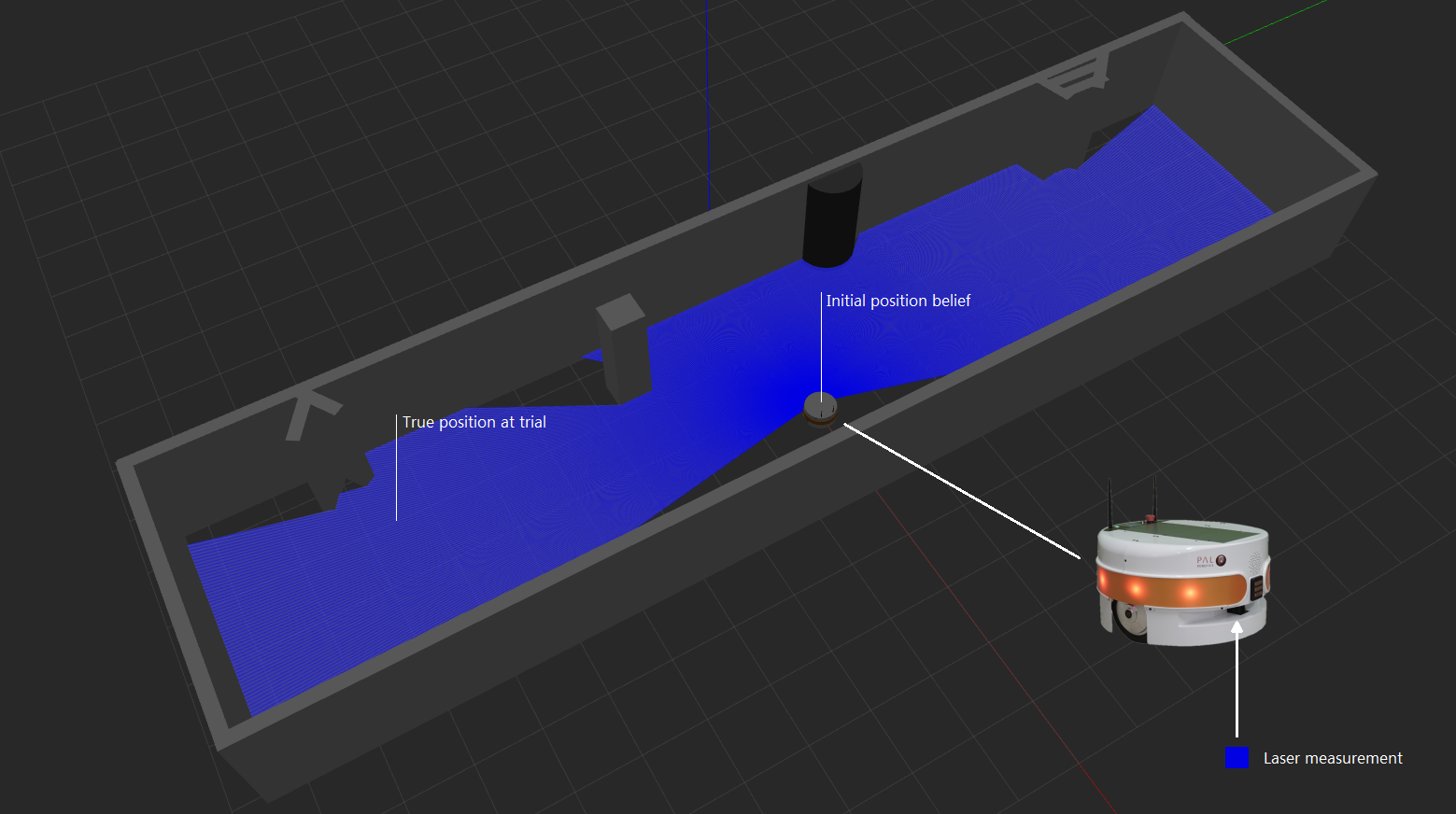}}}
		\caption{Proof-of-concept environment designed for the experiments in Gazebo. In the localization experiment the robot true position is randomly set and the initial position belief is initialized to the center of the map. Laser range finder measurements are displayed as the blue shading.}
		\label{fig:env}
	\end{figure}

	We describe how the predictive processing (PP) approach to perception~\cite{clark2013whatever,lanillos2018adaptive} can aid in localization and simple navigation tasks~\cite{lanillos2021neuroscience}. In this work navigation is performed having the robot move between two points in an unobstructed environment, which can be further built upon to tackle more complex environments (e.g. mazes). Under PP, the agent, following the Free Energy Principle (FEP)~\cite{friston_2010:FEP}, tries to minimize the error in the predicted observations by either performing corrective actions to match the expected internal state or by updating this internal state based on what it has experienced through the senses. In this work, we present a proof-of-concept based on the Pixel-Active Inference model~\cite{pixel_ai}, proposed for humanoid body perception and action, to perform laser-based localization and navigation without the need for odometry. This has been successfully applied to robot manipulator control to improve adaptation~\cite{meo2021multimodal}. Our approach combines the power of deep networks regression with variational Bayesian filtering to provide a better reliable state estimation than PFs in our proof-of-concept environment---See Fig.~\ref{fig:env}.
	
	\section{Methods}
	
	\subsection{Robot} The TIAGo Base mobile robot uses the SICK TiM571 laser sensor, which has a $0.0\text{m}-25\text{m}$ range and a $270^{\circ}$ aperture angle. In all experiments we limited the robot's movement to 2 degrees of freedom, i.e., moving forward, backward and sideways.
	
	\subsection{Localization}
\begin{figure}
		\centerline{\resizebox{\linewidth}{!}{\includegraphics{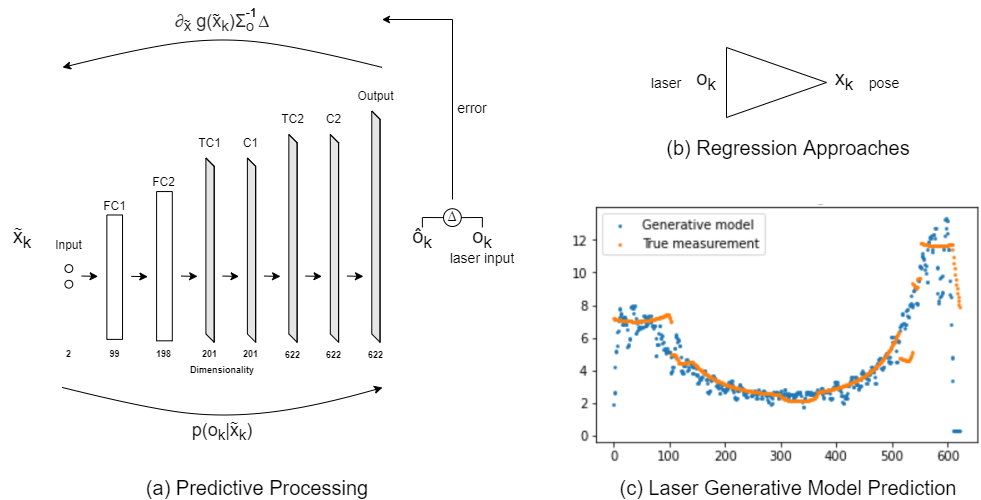}}}
		\caption{(a) The Predictive Processing algorithm's architecture. (b) Graphical representation of regression approaches to robot state estimation. (c) Generative model prediction vs. true laser measurement on a test sample.}
		\label{fig:algo_sketch}
	\end{figure}

	We define the true state (position) of the robot at instant $k$ as $\state_k=(x,y)\in \mathbb{R}^2$ and the position belief of the robot as $\Tilde{\state}_k$. We further define the observation $\obs_k$ as the laser measurements. Estimation is solved by computing the posterior distribution $p(\state|\obs)$ by optimizing the Variational Free Energy (VFE). The algorithm is sketched in Fig. \ref{fig:algo_sketch}a. Under the mean-field and Laplace approximation this is equivalent to minimizing the error between the sensory input $\obs_k$ and the predicted sensory input $\hat{\obs}_k$. While regression approaches (Fig.~\ref{fig:algo_sketch}b) compute the pose directly from the visual input, stochastic neural filtering continuously refines the state through an error signal. We perform state estimation through perceptual inference, minimizing the VFE as follows:
	
	\begin{equation}
	\Tilde{\state} = \argmin_{\Tilde{\state}} F(\Tilde{\state},\obs) \rightarrow \Tilde{\state}_{k+1} = \Tilde{\state}_k + \alpha \partial_{\Tilde{\state}}g(\Tilde{\state}_k)\Sigma_\obs^{-1}(\obs_k - g(\Tilde{\state}_{k}))
	\end{equation}
	
	Where $\alpha$ is the step size and $\partial_{\Tilde{\state}}$ denotes the derivative with respect to $\Tilde{\state}$. This is computed iteratively using gradient descent on the prediction error---sensor measurement $\obs_k$ minus the predicted sensory input $g(\Tilde{\state}_k)$---weighted by the variance $\Sigma_\obs$. Both the predicted observations and the partial derivative of the error are computed by means of a deep neural network forward pass and its Jacobian~\cite{pixel_ai}, respectively.
	
	\subsection{Predicting the observations}
	We compute the sensor likelihood $p(\obs_k|\Tilde{\state}_k)$ using a transposed convolutional neural network (Fig. \ref{fig:algo_sketch}), which augments the dimensionality of the input from $\state$ to the laser-sensor input size (e.g., $2\rightarrow 622$). The input is firstly fed into two fully connected layers, and each transposed convolution layer is followed by a regular convolution layer, based on the work of \cite{pixel_ai}. At every layer, we used the ReLU activation function.
	
	The network was trained on $13000$ normalized random samples collected in the Gazebo simulation. Each sample consists of the true position of the robot and the laser-sensor measurements at that location. The training was performed with $20$ batches of $500$ samples, using the L1 loss and Adam optimizer~\cite{adam_opt}.
	
	\subsection{Navigation}
	Analogously, our algorithm infers the action in the same way that state estimation is computed, namely performing \textit{active inference}. Actions also minimize the VFE in the predicted observations. 
	\begin{equation}
	\action = \argmin_\action F(\Tilde{\state},\obs(\action))
	\end{equation}
	We define the goal as the preference or the intention of the agent $\Tilde{\state}_{goal}$ to arrive to a sensory state $o_{goal}$~\cite{oliver2021empirical}. Estimation and control are computed as follows\footnote{This update equation assumes that the Hessian of the goal dynamics is $-1$ as proposed in \cite{pixel_ai}.}:

	\small
	\begin{align}
	\Tilde{\state}_{k+1} &= \Tilde{\state}_k + \alpha\left[\partial_{\Tilde{\state}}g(\Tilde{\state}_k)\Sigma_\obs^{-1}(\obs_k - g(\Tilde{\state}_{k})) + 
	\partial_{\Tilde{\state}}g(\Tilde{\state}_k)\Sigma_\state^{-1}\beta(\obs_k - g(\Tilde{\state}_{goal}))
	\right]\label{eq:perception}\\
	\action_{k+1} &= \action_k + \gamma\partial_a \Tilde{\state} \partial_{\Tilde{\state}}g(\Tilde{\state}_k)\Sigma_\obs^{-1}(\obs_k - g(\Tilde{\state}_{k}))
	\label{eq:action}
	\end{align}
	\normalsize
	where $\beta$ weights the goal attractor and $\gamma$ is the action step size. Note that each term computes the weighted prediction error mapped to the latent space.
	
	The estimated state $\Tilde{\state}$, now biased by the desired goal, generates a new predicted observation at every new iteration that is transformed into an action $\action$, which minimizes the VFE. Thus, performing a movement in the direction of the goal. The pseudo-code described in Alg. \ref{alg:algorithm} illustrates the process. The algorithm converges when the observation fits the predicted laser sensor measurements.
	
	\begin{algorithm}[hbtp!]
		\SetAlgoLined
		$\Tilde{\state} \leftarrow$ initial belief\;
		$\obs_{goal} \leftarrow g(\Tilde{\state}_{goal})$\tcp*[l]{Generate goal}
		\While{true}{
			$\obs_k \leftarrow \text{Normalize(laser input)}$\;
			$\hat{\obs}_k \leftarrow g(\Tilde{\state})$\tcp*[l]{Predicted observation}
			$\Tilde{\state} \leftarrow \text{Eq. \ref{eq:perception}}$\;
			$\action \leftarrow \text{Eq. \ref{eq:action}}$\;
			PerformAction($a$)\;
		}
		\caption{FEP localization and navigation algorithm}
		\label{alg:algorithm}
	\end{algorithm}

	
	\section{Results}
\begin{figure}
		\centerline{\resizebox{10cm}{!}{\includegraphics{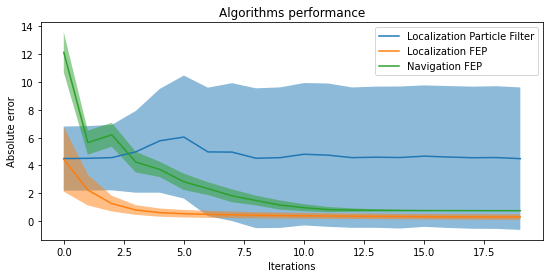}}}
		\caption{Localization and navigation evaluation. Mean and standard deviation of the positioning absolute error of our model (FEP) compared against the PF algorithm. Furthermore, our model is able to perform navigation using the same Bayesian filtering framework. The green line shows the mean absolute error to the goal.}
		\label{fig:results}
	\end{figure}

	We evaluated our laser-based Active Inference algorithm against a particle filter~\cite{particle_filter} in the Gazebo simulator, using a commercial mobile robot with a laser rangefinder sensor (TIAGo Base, pmb-2)~\cite{tiago_base}, interfaced with Robot Operating System (ROS)~\cite{quigley2009ros}. All experiments were conducted in a designed corridor-like map described in Fig. \ref{fig:env}. Localization and navigation results are summarized in Fig.~\ref{fig:results}.
	
	\subsection{Localization and estimation}
	Firstly, we evaluated the localization accuracy when initializing the robot to a random position in the space when a map was given. Thus, testing absolute positing with laser measurements in static situations. For the particle filter, the particle initial probabilities were randomly spread in the environment and reset at the beginning of every trial. In our algorithm, we initialized the initial belief in the center of the environment. We computed the ground truth positional error in 100 trials for 50 iterations. Our algorithm converged much more consistently to the true state over all trials, whereas the PF struggled to deliver consistent results, as shown in  Fig. \ref{fig:results} by the rather large standard deviation in the blue shaded area. It is important to highlight that the PF is tuned for using the robot's odometry. However, for the sake of fair comparison solely laser-sensor values were used as observations.
	
	Secondly, we evaluated the localization performance when traversing the environment from one side to the other, by performing small teleports (to override odometry) to simulate robot movement while keeping the rotation angle constant. Results showed a more stable over time state estimation by our algorithm when compared to the PF, which seemed to suffer from the absence of odometry information.
	
	\subsection{Navigation}
	For the assessment of the navigation algorithm's performance, we ran an experiment consisting of 50 trials in which the robot had to navigate from a starting point to a goal position. The initial belief state was set to the robot's initial true position, to evaluate the performance of navigation without the effects of localization in the first iterations. In every trial, both the initial position and the goal state of the robot were chosen randomly, with the constraint that they should be at least 12 meters (in Gazebo coordinates) apart from each other. The task was considered complete when the robot got in a range of 0.8m from the target. The results are plotted in green in Fig.~\ref{fig:results}.
	
	We observed that the robot initially quickly approximates the goal, with a big drop in the distance to the goal in the first couple of iterations. As it gets closer to the goal, the ``velocity" of the robot (in the experiment described by the step sizes) decreases. This is a result of the diminishing gradient in every step of the algorithm, due to the stochastic gradient descent. Additionally, we computed the average number of iterations that it took the algorithm to get in the desired $0.8$m range of the target. Over the 50 trials of similar travel distance ($\sim{11m}$ to $\sim{13.5m}$), the average number of iterations was $12.5$. This number is naturally closely related to the optimal step size found.

	\section{Conclusions}\label{conclusions}
	This work shows a proof-of-concept on how predictive processing, i.e. active inference agents, can perform laser-based localization and navigation tasks. The results obtained in the localization experiment, where we compared our approach against a state-of-the-art alternative (particle filter), show the potential of predictive stochastic neural filtering in robot localization, and estimation in general~\cite{friston2008variational,millidge2021neural}. Furthermore, the navigation experiment showcased how to compute actions as a dual filtering process. Nevertheless, at its current state, the proposed algorithm suffers from a few deficiencies, most of which are related to the learning of the generative model of the world. Besides currently requiring a large dataset for training, the model is prone to mistake very similar objects in the environment, given that the estimation of the new state is independent from the previous. Additionally, because it is a supervised method trained before that the robot can do any navigation, it is unable to cope with changing environments. All in all, the environment used in our experiments is rather simplistic compared to demonstrations of current sota algorithms. Therefore, we foresee further development and experimentation in terms of integration of odometry information, the introduction of extra degrees of freedom and connection to the robot's non-linear dynamics.

	\bibliographystyle{splncs04}
	\bibliography{bibliography}
	
\end{document}